\documentclass[10pt,twocolumn,letterpaper]{article}

\usepackage[pagenumbers]{cvpr}              %

\usepackage{textcomp}
\usepackage{colortbl}
\usepackage[table]{xcolor}
\usepackage{xcolor}
\usepackage{pifont}
\newcommand{\cmark}{\textcolor{ForestGreen}{\ding{51}}} %
\newcommand{\xmark}{\textcolor{BrickRed}{\ding{55}}}    %

\newcommand{\dmark}{\textcolor{gray}{$\diamond$}}

\definecolor{cvprblue}{rgb}{0.21,0.49,0.74}
\usepackage[pagebackref,breaklinks,colorlinks,allcolors=cvprblue]{hyperref}
\usepackage[capitalize,noabbrev]{cleveref}
\usepackage{booktabs,adjustbox,multirow,multicol}
\usepackage{xcolor}
\def\paperID{20} %
\def\confName{CVPR}
\def\confYear{2026}

\title{VeRVE: Versatile Retrieval for Videos via Unified Embeddings}

\author{Shaunak Halbe\\
Georgia Institute of Technology\\
{\tt\small shalbe9@gatech.edu}
\and
Bhagyashree Puranik\\
Amazon\\
{\tt\small bpuranik@amazon.com}
\and
Jayakrishnan Unnikrishnan\\
Amazon\\
{\tt\small jayunn@amazon.com}
\and
Kushan Thakkar\\
Amazon\\
{\tt\small tkushan@amazon.com}
\and
Vimal Bhat\\
Amazon\\
{\tt\small vimalb@amazon.com}
\and
Toufiq Parag\\
Keystone AI\\
{\tt\small toufiq.parag@gmail.com}
}

\begin{document}
\maketitle

\begin{abstract}
Modern video retrieval systems are expected to handle diverse tasks ranging from corpus-level retrieval, fine-grained moment localization to flexible multimodal querying. Specialized architectures achieve strong retrieval performance by training modality-specific encoders on massive datasets, but they lack the ability to process composed multimodal queries. In contrast, multimodal LLM (MLLM)-based methods support rich multimodal search but their retrieval performance remains well below that of specialized systems. We present \textit{VeRVE}, an MLLM-based versatile video retrieval framework that integrates corpus and moment-level retrieval capabilities while accommodating composed multimodal queries within a single architecture. We use contrastive alignment of visual and textual embeddings generated using a shared MLLM backbone to facilitate efficient embedding-based candidate search. Our embedding model, trained efficiently using low-rank adaptation (LoRA) on $700$K paired visual-text data samples, surpasses other MLLM-based methods on zero-shot video retrieval tasks. Additionally, we demonstrate that the same model can be adapted without further training to achieve competitive results on zero-shot moment retrieval, and state of the art results for zero-shot composed video retrieval. With additional training for reranking candidates identified in the embedding-based search, our model substantially outperforms existing MLLM-based retrieval systems and achieves retrieval performance comparable to state of the art specialized models.
\end{abstract}
     
\section{Introduction}
\label{sec:intro}
\begin{figure*}[t]
\centering
\begin{minipage}[c]{0.42\textwidth}
\centering
\begingroup
\small
\setlength{\tabcolsep}{3pt}
\renewcommand{\arraystretch}{1.05}

\scalebox{0.95}{
\begin{tabular}{lcccc}
\toprule
\rowcolor{gray!20}
\textbf{Method} &
\textbf{Unified} &
\textbf{Re-} &
\textbf{Moment} &
\textbf{Composed} \\
\rowcolor{gray!20}
& \textbf{Embed.} & \textbf{ranking} & \textbf{Retr.} & \textbf{Video Retr.} \\
\midrule
InternVideo2          & \xmark & \cmark & \xmark & \xmark \\
LamRA                 & \cmark & \cmark & \xmark & \dmark \\
CaRe                  & \cmark & \xmark & \xmark & \dmark \\
Chat-VTG              & \xmark & \xmark & \cmark & \xmark \\
\textit{VeRVE} & \cmark & \cmark & \cmark & \cmark \\
\bottomrule
\end{tabular}
}

\vspace{2mm}

\endgroup
\end{minipage}
\hfill
\begin{minipage}[c]{0.57\textwidth}
\centering
\includegraphics[width=0.99\textwidth]{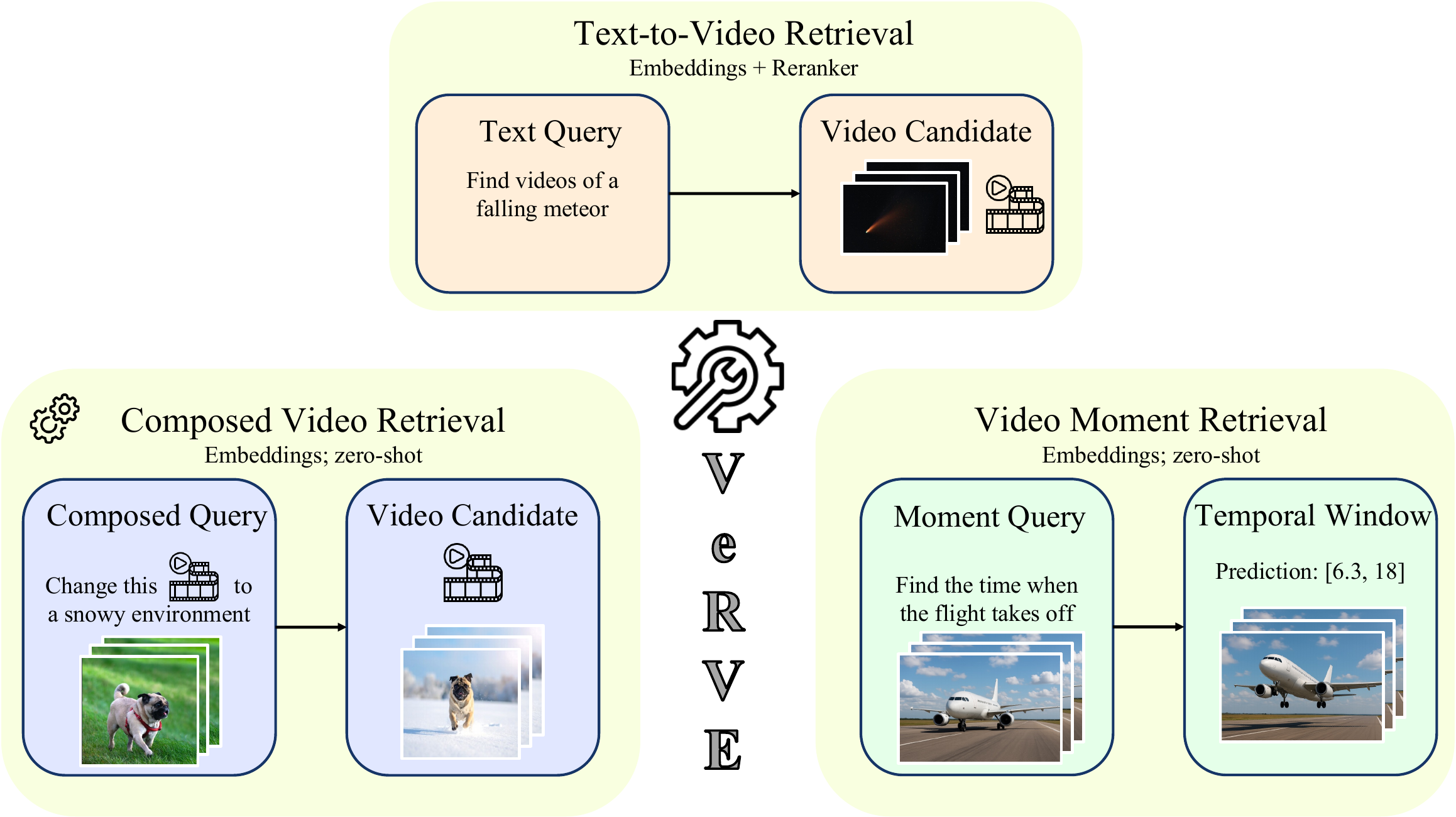}
\end{minipage}
\caption{\textit{VeRVE} supports corpus-level retrieval with reranking, zero-shot composed video retrieval, and zero-shot moment localization within a single architecture. The table highlights that \textit{VeRVE} uniquely offers unified embeddings and versatile capabilities. {\dmark } indicates models that, while architecturally capable of processing multimodal inputs, have not demonstrated composed video retrieval capability.}
\label{fig:overview}
\end{figure*}

Video content constitutes the majority of internet data today, creating a demand for effective and scalable search strategies that enable users to discover relevant content. As video collections continue to expand in size and diversity, retrieval systems must accommodate increasingly complex and fine-grained user specifications. It is desirable that an effective video retrieval framework handle diverse retrieval scenarios, such as (a) \textbf{corpus-level retrieval}, which identifies relevant videos from a large collection via dense embedding search followed by an optional re-ranking stage; (b) \textbf{composed multimodal querying}, where users can issue queries combining multiple modalities—text, images, or videos—to express complex intent (e.g., “find videos like \emph{$\langle$this clip$\rangle$}, but set in a snowy environment” as in \cref{fig:overview}); and (c) \textbf{moment retrieval}, which localizes the precise temporal segment within a video that matches a natural-language query (e.g., ``the flight takes off” as in \cref{fig:overview}). These capabilities form the foundation for intelligent, flexible, and user-centric video search and recommendation systems \cite{Ventura_2024,hummel2024egocvregocentricbenchmarkfinegrained}.

Early video retrieval models typically employed modality-specific encoders with intermediate or late fusion to align video and text representations~\cite{bain2022frozentimejointvideo,bain2022cliphitchhikersguidelongvideo,luo2021clip4clipempiricalstudyclip}. These encoders produce embeddings for different modalities in a shared embedding space which enables cross-modal retrieval. Specialized architectures such as InternVideo2~\cite{wang2024internvideo2scalingfoundationmodels} push this paradigm to scale, training on hundreds of millions of paired examples across video, image and audio, to achieve strong text-to-video retrieval performance. However, since these systems rely on separate modality-specific encoders, they cannot natively process composed queries that combine multiple input modalities.

Multimodal large language models (MLLMs) have gained significant attention due to their ability to jointly process visual and textual information~\cite{li2023blip2bootstrappinglanguageimagepretraining, bai2025qwen25vltechnicalreport,zhang2025llavavideovideoinstructiontuning,liu2023visualinstructiontuning}.  %
Recent efforts have adapted MLLMs for retrieval tasks via contrastive learning to produce unified embeddings~\cite{liu2024lamralargemultimodalmodel,zhang2025gmeimprovinguniversalmultimodal,jiang2025vlm2vectrainingvisionlanguagemodels,jiang2024e5vuniversalembeddingsmultimodal}. A common recipe for adapting these models involves training on diverse data mixtures spanning a variety of tasks such as visual question answering, captioning, retrieval and grounding. Some works further incorporate an instruction tuning stage~\cite{liu2024lamralargemultimodalmodel, xu2025carebenchfinegrainedbenchmarkvideo} to improve vision-language alignment. These approaches prioritize generalization, aiming to enable flexible retrieval across many modalities including images, text, and multimodal documents. However, while extensible to video retrieval, this broad focus comes at a cost: such models considerably underperform specialized approaches like InternVideo2 on standard corpus-level video retrieval benchmarks \cite{7780940,hendricks2017localizingmomentsvideonatural}. %

In parallel, MLLMs have also been adapted for moment retrieval. Some recent approaches fine-tune MLLMs with temporal annotation data ~\cite{lu2024llavamrlargelanguageandvisionassistant,Meinardus2024TheSurprisingEffectivenessofMultimodal}, while others generate video captions to obtain temporal proposals~\cite{qu2024chatvtgvideotemporalgrounding} or introduce architectural modifications in the form of temporal processing modules ~\cite{qian2024momentoradvancingvideolarge}. However these two lines of work remain disconnected. To our knowledge, no single architecture simultaneously delivers strong corpus-level video retrieval, performs zero-shot composed video retrieval, and enables zero-shot moment localization without sacrificing performance compared to task-specific models.

In this paper, we propose \textit{VeRVE}, a unified MLLM-based video retrieval framework that jointly supports corpus-level retrieval with reranking, composed multimodal querying, and moment localization. Unlike prior multi-objective or instruction-tuned methods~\cite{liu2024lamralargemultimodalmodel,zhang2025gmeimprovinguniversalmultimodal,jiang2025vlm2vectrainingvisionlanguagemodels}, our approach follows a retrieval-centric design centered on contrastive alignment between visual and textual inputs. 
We repurpose the final hidden state corresponding to the MLLM’s end-of-sequence (\verb|EOS|) token as a unified embedding for both queries and videos through contrastive alignment. We fine-tune low-rank adapters to produce embeddings and term this model \textit{VeRVE-Embed}. We subsequently train a re-ranker (termed {\textit{VeRVE-Ranker}}), relying upon the same backbone and including a linear projection head that predicts a matching score between $[0, 1]$ (as opposed to generating ``Yes/No" responses~\cite{liu2024lamralargemultimodalmodel}) for query–video pairs. %
Despite its simplicity, our framework generalizes across tasks without architectural changes or task-specific supervision.
Importantly, the contrastively aligned embeddings themselves (before the reranker) serve as effective retrieval representations, achieving competitive performance relative to prior embedding models. Relying only on simple vision–text training, we achieve competitive results approaching the accuracy of methods trained using additional modalities like audio \cite{wang2024internvideo2scalingfoundationmodels, cicchetti2025gramianmultimodalrepresentationlearning}. To extend our method to moment retrieval, we propose a simple yet effective pipeline built on top of our \textit{VeRVE-Embed} model, enabling the prediction of temporal windows corresponding to natural language queries. Crucially, our pipeline predicts variable-sized windows in a single step without the multi-stage refinements required by existing works \cite{qu2024chatvtgvideotemporalgrounding}, while remaining highly computationally efficient. 

Empirically, \textit{VeRVE} achieves text-to-video retrieval performance on par with specialized architectures, while substantially outperforming prior MLLM-based methods on MSR-VTT~\cite{7780940}, DiDeMo~\cite{hendricks2017localizingmomentsvideonatural} and MSVD~\cite{chen-dolan-2011-collecting} test sets. Notably, our model even surpasses the state of the art (SOTA) approach ~\cite{wang2024internvideo2scalingfoundationmodels} on DiDeMo~\cite{hendricks2017localizingmomentsvideonatural} for zero-shot text-to-video retrieval. It further achieves SOTA zero-shot composed video retrieval on CoVR~\cite{Ventura_2024} test set, strongly surpassing the best performing approach~\cite{thawakar2024composedvideoretrievalenriched}, and demonstrates strong zero-shot moment retrieval performance on Charades-STA~\cite{gao2017talltemporalactivitylocalization} and ActivityNet-Captions ~\cite{krishna2017densecaptioningeventsvideos}.

Our main contributions are as follows:  
(i) we introduce \textit{VeRVE}, a versatile MLLM-based framework that handles corpus-level retrieval with reranking, fine-grained moment localization, and composed multimodal querying within a single architecture;
(ii) \textit{VeRVE} achieves text-to-video retrieval performance comparable to specialized systems such as InternVideo2, despite not training on audio, while supporting composed queries that such systems cannot process;
(iii) our contrastively trained embedding model demonstrates strong emergent capability on unseen tasks,  achieving state of the art results on zero-shot composed video retrieval (CoVR) and strong performance on zero-shot moment retrieval through our effective temporal localization pipeline.

\section{Related Work}
\label{sec:rw}

\subsection{Corpus-level video retrieval}

Prior works on text-to-video and video-to-text retrieval typically employ dual-encoder approaches \cite{miech2019howto100mlearningtextvideoembedding, miech2020endtoendlearningvisualrepresentations} trained using contrastive learning on text-video pairs. Query embeddings generated by the encoders are used for dense vector search followed by an optional re-ranking stage. Methods  such as CLIP4Clip \cite{luo2021clip4clipempiricalstudyclip} and others \cite{bain2022frozentimejointvideo,bain2022cliphitchhikersguidelongvideo}  extend CLIP~\cite{radford2021learningtransferablevisualmodels} by transferring image-text alignment to video by applying frame-level encoders followed by late-fusion aggregation, but still preserve separate modality streams. Recent works like VidLA \cite{rizve2024vidlavideolanguagealignmentscale}, GRAM \cite{cicchetti2025gramianmultimodalrepresentationlearning} and InternVideo2 \cite{wang2024internvideo2scalingfoundationmodels} pursue stronger multimodal alignment and large-scale training strongly boosting benchmark performance in both fine-tuned and zero-shot settings. 
A variant of traditional video retrieval is the recently proposed composed video retrieval task~\cite{Ventura_2024} in which the query is in the form of a video along with a desired modification. Prior solutions use adaptations of the BLIP~\cite{li2022blipbootstrappinglanguageimagepretraining} model to tackle this task.

\subsection{Multimodal Large Language Models}

MLLMs connect vision encoders to large language models through lightweight adapters \cite{alayrac2022flamingovisuallanguagemodel, li2023blip2bootstrappinglanguageimagepretraining, liu2023visualinstructiontuning} or visual expert modules \cite{wang2024cogvlmvisualexpertpretrained, chen2024internvl}, achieving strong performance on a broad range of vision-language tasks. Recent models like Qwen2.5-VL \cite{bai2025qwen25vltechnicalreport} demonstrate enhanced multimodal reasoning with native video processing capabilities for extended sequences. These advances have driven widespread adoption of MLLMs across diverse applications, particularly in video understanding tasks where their ability to process temporal context and composed queries opens new possibilities for video-language modeling.
\begin{figure*}[t]
    \centering
    \vspace{-0.5cm}\includegraphics[width=0.9\textwidth]{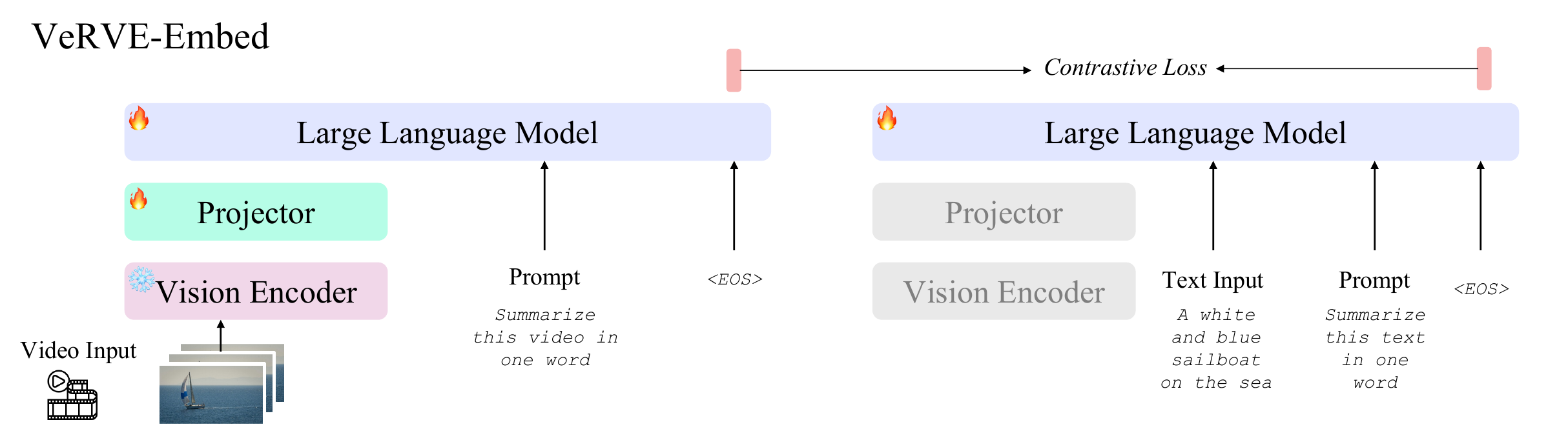}
    \caption{\textit{VeRVE-Embed} uses the final hidden state of the EOS token as an embedding anchor, and aligns visual content and text descriptions through contrastive learning.\vspace{-0.4cm}} 
    \label{fig:corpus}
    
\end{figure*}

\subsection{MLLMs as unified embedding models}

As MLLMs are primarily designed for generation and question-answering, they cannot natively generate embeddings for retrieval. Nevertheless their unified representation space makes them natural candidates for retrieval applications. Recent works have explored adapting MLLMs as unified embedding models through contrastive learning. E5-V \cite{jiang2024e5vuniversalembeddingsmultimodal} demonstrates this potential by fine-tuning LLaVA on text-only natural language inference (NLI) data for cross-modal retrieval. GME \cite{zhang2025gmeimprovinguniversalmultimodal} achieves stronger performance by training Qwen-2VL on a mixture of image-text data, while subsequent approaches such as LaMRA \cite{liu2024lamralargemultimodalmodel}, VLM2Vec \cite{jiang2025vlm2vectrainingvisionlanguagemodels}, and CaRe \cite{xu2025carebenchfinegrainedbenchmarkvideo} improve on this through refined data mixtures, modality-specific curation, and hard negative mining strategies. However, these methods predominantly target image and document retrieval, with limited exploration of video-specific capabilities. 

\subsection{MLLMs for moment retrieval}
Recent MLLM-based works address video moment retrieval with varying degrees of specialization. TimeChat~\cite{ren2024timechattimesensitivemultimodallarge} achieves zero-shot temporal grounding through instruction tuning on diverse temporal tasks with timestamp-aware encoding. Training-free approaches like ChatVTG~\cite{qu2024chatvtgvideotemporalgrounding} rely on external MLLMs to generate multi-perspective captions for semantic matching. HawkEye~\cite{wang2024hawkeyetrainingvideotextllms} and Momenter~\cite{qian2024momentoradvancingvideolarge} construct large-scale datasets with segment-level temporal annotations to enable text-to-text grounding and fine-grained temporal reasoning. In contrast, we demonstrate our simple, zero-shot pipeline that matches the performance of these specialized approaches without requiring task-specific modifications, external captioning models, or query reformulation.

While prior methods have bespoke MLLM-based solutions for various retrieval tasks, in this work we address the full spectrum of video retrieval, including corpus-level search, moment-level localization and composed video retrieval.

\section{Method}
\label{sec:method}
We propose \textit{VeRVE} (\textbf{Ve}rsatile \textbf{R}etrieval for \textbf{V}ideos via unified \textbf{E}mbeddings), a versatile MLLM-based framework for video retrieval that addresses multiple tasks including corpus-level retrieval, moment localization, and composed video retrieval without needing task-specific modifications. We adapt a MLLM backbone to produce unified embeddings for retrieval through contrastive learning. As detailed in Section~\ref{sec:corpus}, we adopt a contrastive training scheme on relatively small datasets to effectively learn multimodal representations for video retrieval. This approach is simpler than the training recipes on multi-task data mixtures employed by prior methods~\cite{jiang2025vlm2vectrainingvisionlanguagemodels,liu2024lamralargemultimodalmodel}. 
Our training strategy first adapts the MLLM with image-text data, followed by video-text training. This design is motivated by recent studies demonstrating that semantic representations learned from images provide strong priors for video understanding~\cite{xu2024pllavaparameterfreellava,shen2024longvuspatiotemporaladaptivecompression}. This yields our \textit{VeRVE-Embed} model, which we use to obtain candidate videos for corpus-level retrieval, enable composed video retrieval, and perform zero-shot moment localization. For corpus-level retrieval, we refine the initial candidates obtained from \textit{VeRVE-Embed} through a re-ranking stage, which we term \textit{VeRVE-Ranker}. In contrast to existing strategies~\cite{liu2024lamralargemultimodalmodel}, we formulate our re-ranker as a video-query matching problem and train a classification head on the EOS output embedding from the MLLM to predict a confidence score in range of $[0,1]$ for each video-query pair. Section~\ref{S:RERANKING} details our novel re-ranking objective along with the different negative sampling strategies we employ. %
Beyond corpus-level video retrieval, \textit{VeRVE-Embed} directly supports moment localization (Section~\ref{S:MOMENT}) and composed video retrieval (Section~\ref{S:composed}) in a zero-shot manner, requiring neither additional architectural modifications nor task-specific fine-tuning.

\subsection{{VeRVE-Embed}: Contrastive Visual Embedding Learning}
\label{sec:corpus}
MLLMs are typically trained for autoregressive text generation. We adapt them for producing representation embeddings by utilizing the final hidden state of the end-of-sequence (\verb|EOS|) token as an embedding anchor. The causal attention mechanism in the LLM ensures that this token attends to the full multimodal context, capturing a global summary of the input.

\textbf{Contrastive objective.} We use the same model to encode both candidates and text queries in separate forward passes, as illustrated in Figure~\ref{fig:corpus}. For each modality, we append a prompt instruction to summarize the content into one word, followed by the \verb|EOS| token. Let $q_i$ denote the embedding of a query and $c_i$ denote the embedding of its paired candidate. We train using the InfoNCE loss \cite{oord2019representationlearningcontrastivepredictive} to train the model as follows:
\vspace{-0.3cm}
\begin{equation}  \mathcal{L}_{\text{InfoNCE}} = -\log \frac{\exp(\text{sim}(q_i, c_i)/\tau)}{\sum_{j=1}^{N} \exp(\text{sim}(q_i, c_j)/\tau)} \vspace{-0.2cm}
\end{equation}
where $\text{sim}(\cdot, \cdot)$ denotes cosine similarity, $\tau$ is the temperature parameter, and $N$ is the batch size. For efficient adaptation, we use LoRA~\cite{hu2021loralowrankadaptationlarge} adapters within attention and projection layers while keeping the base model weights frozen.
 
\textbf{Training recipe.} We first train the MLLM backbone on image-caption pairs to establish vision-language alignment, then fine-tune on video-text description pairs. The image-text dataset (CC-595K) provides broad concept coverage and enables efficient vision-language alignment. We then fine-tune on a relatively small video-text dataset (PEVideo) to adapt the model to video data. For image-text data, the visual input consists of a single image, while for video-text pairs, we use uniformly sample frames from the video. Through this contrastive alignment of \verb|EOS| embeddings across modalities, the model learns a unified representation space suitable for multimodal video retrieval.

\subsection{{VeRVE-Ranker}: Candidate Refinement}
\label{S:RERANKING}
\begin{figure}[t]
    \centering
    \centering
    \vspace{-0.3cm}\includegraphics[width=0.4\textwidth]{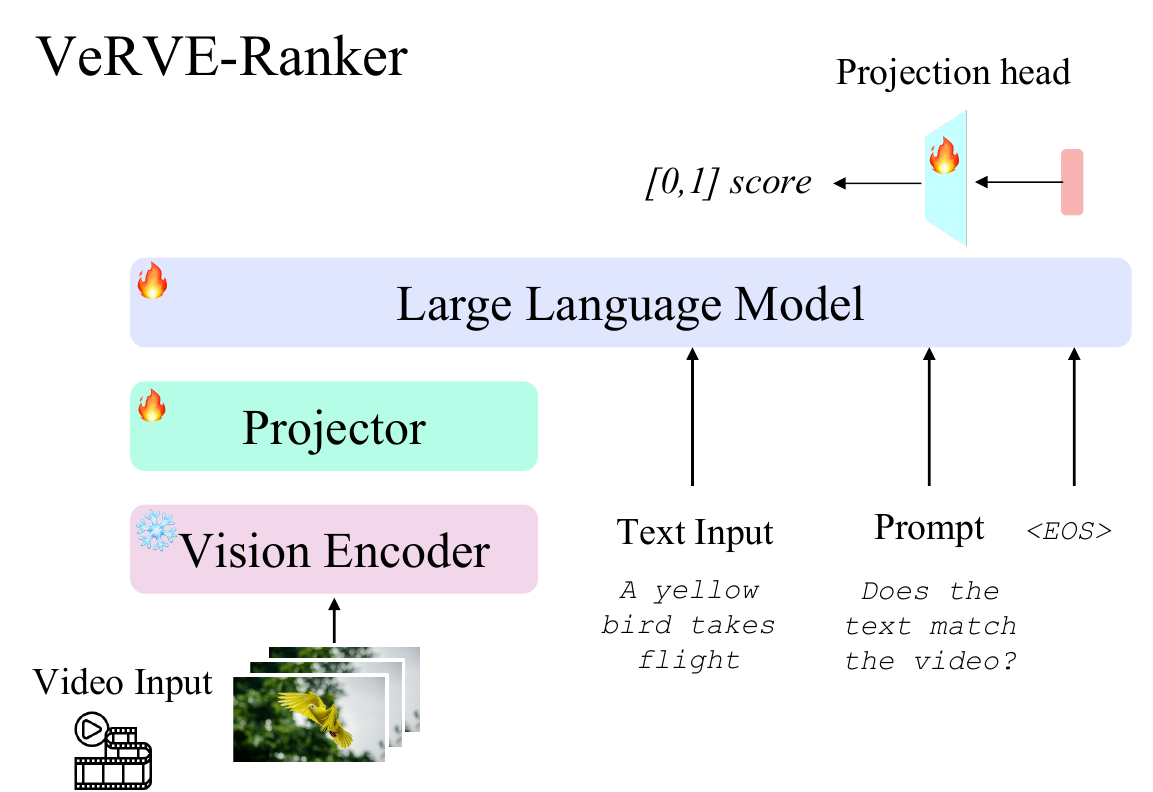}
    \caption{\textit{VeRVE-Ranker} re-scores each query–video pair by feeding them jointly through the MLLM and projecting the EOS hidden state to a pointwise matching score.}
    \label{fig:reranking}
    \vspace{-5mm}
\end{figure}

Once trained with the image/video to text contrastive objective, \textit{VeRVE-Embed} can be used at inference time to retrieve an initial set of candidates by computing cosine similarities between the query embedding and all video embeddings in the corpus. While this embedding-based retrieval enables efficient search over large collections, it relies on fixed-dimensional representations that may not fully capture the semantic complexity of video content. To address this, we introduce a re-ranking stage that operates on the top-$K$ retrieved candidates. Unlike the initial retrieval, which requires scalability to millions of videos, re-ranking processes a small candidate set, enabling fine-grained cross-modal interaction.

We formulate re-ranking as a pointwise binary matching problem between a query $q$ and each candidate $c$. Each query-candidate pair is jointly processed by the MLLM. We initialize a linear projection head and pass the final hidden state of the \verb|EOS| token through it to predict a confidence score $s_\theta(q,c)$ in $[0,1]$, where $1$ indicates a correct match. During training, we optimize both the linear projection head and the LoRA adapters in LLM self-attention and projection layers, while keeping the base model weights frozen. We train to re-rank exclusively with video-text pairs. At inference, candidates are re-ranked according to their predicted confidence scores.

\noindent \textbf{Training objective.}
Training the re-ranker requires both positive (matching) and negative (non-matching) query-candidate pairs. We utilize the following sampling strategies to source negatives:
\begin{itemize}
    \item \textbf{Random negative sampling:} For each query, we randomly sample non-matching targets from the training set. 
    \item \textbf{Hard negative mining:}  For each query, we first retrieve the top-$K$ candidates using our \textit{VeRVE-Embed} model and sample non-matching targets from this set. Since these hard negatives share semantic similarity with the query, this encourages the model to learn fine-grained discriminative features. %
\end{itemize}
To provide an effective training signal, we leverage both random and hard negatives through a joint objective. We first apply BCE loss to both random and hard negatives with binary labels. 
However, BCE treats all negatives equally with a fixed label of 0, which can be suboptimal when hard negatives exhibit partial semantic alignment with the query and forcing the model to assign them scores of 0 may conflict with their actual relevance. To address this, we augment the BCE loss with a preference-based objective following the Bradley-Terry formulation~\cite{bradley1952RankAnalysisofIncompleteBlockDesigns}. Rather than enforcing absolute relevance scores, we optimize for correct relative ordering between ground-truth and negative pairs. For a given query $q$, ground-truth target $c_{gt}$, and negative candidate $c_{neg}$, the loss can be stated as:
\vspace{-0.2cm}
\begin{equation}
    \mathcal{L}_{\rm PB} = -\log \sigma\!\bigl(s_\theta(q,c_{\rm gt}) - s_\theta(q,c_{\rm neg})\bigr)\vspace{-2mm}
\end{equation}

This objective directly follows from the aim of preferring the ground truth target to the mined negative. It encourages $s_\theta(q,c_{\rm gt}) > s_\theta(q,c_{\rm neg})$ enabling the model to rank partially relevant negatives below ground-truth matches without forcing on absolute relevance scores. The complete joint objective becomes: \vspace{-0.3cm}

\vspace{-1mm}
\begin{equation}
    \mathcal{L} = \lambda_1 \mathcal{L}_{\text{BCE-rand}} + \lambda_2 \mathcal{L}_{\text{BCE-hard}} + \lambda_3 \mathcal{L}_{\text{PB}},
\end{equation}
where $\lambda_i$ control relative contribution of each component, and $\mathcal{L}_{\text{BCE-rand}}$, $\mathcal{L}_{\text{BCE-hard}}$ are BCE losses computed on (query, negative) pairs, where negatives are randomly sampled or mined respectively. $\mathcal{L}_{\text{PB}}$ operates on (query, ground-truth target, mined negative) triplets to enforce correct relative ordering. This joint formulation allows a more fine-grained precision by leveraging complementary supervision signals.

\vspace{0mm}
\subsection{Zero-shot Moment Retrieval}\label{S:MOMENT}
For moment retrieval, the objective is to localize the temporal segment (start and end timestamps) within a video that corresponds to a natural language query. Unlike corpus-level retrieval which identifies relevant videos from a large collection, moment retrieval requires fine-grained temporal localization within a single video. We approach this task in a zero-shot manner, leveraging only the contrastively trained embedding model (\textit{VeRVE-Embed}), without any moment-specific training or architectural modifications.

To predict temporal segments relevant to a query, we begin by computing frame-level similarity scores between the query and sampled video frames. Specifically, we uniformly sample frames from the video and encode each frame individually using \textit{VeRVE-Embed}. The query text is encoded once, and we compute cosine similarity between the query embedding and each frame embedding, producing a temporal similarity signal
 $s(t)$ where $t$ is the frame index. Because this signal is computed independently over a discrete set of frames, it is susceptible to abrupt local fluctuations caused by uninformative frames or transient visual artifacts. We apply Gaussian smoothing to $s(t)$ to suppress these spurious local maxima. This step leverages the inherent temporal continuity of video, as relevant semantic events span contiguous frames rather than isolated ones, and produces a coherent, refined signal $\tilde{s}(t)$. Since this task expects contiguous temporal segments as outputs, we first identify frame indices where the similarity scores attain a peak value. Peaks are detected based on a threshold defined as $\mu + \beta \sigma$, where $\mu$ and $\sigma$ are mean and standard deviation of our smoothed signal $\tilde{s}(t)$. For each detected peak at frame $t_p$, we determine the temporal extent by expanding left and right until the similarity falls below $\tilde{s}(t_p) - (1-\alpha)(\tilde{s}(t_p) - \mu)$ to produce candidate windows, where $\alpha$ controls the window tightness around peaks. Neighboring frames can often yield redundant candidate windows. We therefore apply temporal Non-Maximum Suppression (NMS) to eliminate overlaps and retain only the highest scoring segments as our final predictions. This pipeline enables dynamic window predictions in a fully zero-shot manner. 

\subsection{Zero-shot Composed Video Retrieval}\label{S:composed}
Although \textit{VeRVE-Embed} is trained primarily on video and paired text-only data, the underlying MLLM supports encoding multimodal inputs, including text combined with video as an interleaved sequence. The ability to query composed modalities emerges from the MLLM’s joint representation space and multimodal reasoning abilities, enabling retrieval scenarios not explicitly encountered during training. We address this task in a purely zero-shot regime without exposure to any composed multimodal training examples, demonstrating our model's ability to generalize to complex multimodal queries. To process composed queries, we construct the input by concatenating: (1) source video, (2) modification text, (3) an instruction prompting the model to encode the semantic change the source video would undergo under the modification, and (4) the \verb|EOS| token. The resulting embedding serves as the composed query representation. We then retrieve the target video by computing argmax over cosine similarities between this query embedding and all video embeddings in the corpus.

\section{Experimental Analysis}

\subsection{Text-to-Video Retrieval}

\subsubsection{Training Details}

\textbf{Contrastive Learning:} We use the 7B parameter variant of Qwen-VL 2.5 as the backbone and finetune LoRA adapters on its projection layers and LLM blocks with the number of trainable parameters being approximately $48$M. The corpus-level training uses an image-text followed by video-text alignment setup designed to learn cross-modal representations. First, the model is trained on image-text pairs from CC 595K, a balanced subset of Conceptual Captions 3M \cite{sharma-etal-2018-conceptual}. Next, we continue training on $105$K video-text pairs from PEVideo \cite{bolya2025perceptionencoderbestvisual}, focusing on temporal alignment between short video clips and their corresponding captions. The videos in PEVideo are sourced from the Segment Anything corpus, consisting of real-world, in-the-wild video clips. Starting with image-text alignment allows broad concept coverage, enabling the model to learn general-purpose visual-text alignment far more efficiently compared to solely training on a larger video-text dataset. For the image dataset we train for $3$ epochs, and then continue on the video data for $1$ epoch. We use gradient accumulation with a step size of $4$ and a per-device batch size of $64$ for images and $8$ for videos, resulting in effective global batch sizes of $2048$ and $256$ respectively. 

\textbf{Re-ranking:} For candidate re-ranking, we train the same backbone on PEVideo using the training strategy discussed in Section~\ref{S:RERANKING} for $1$ epoch. Per training example, we sample $1$ random negative, $1$ mined negative for $\text{BCE-hard}$ and $1$ mined negative for the preference-based objective. To obtain the mined negatives, we retrieve the top-$50$ candidates from the dataset per query using the corpus-level model (\textit{VeRVE-Embed}) and sample from the range $[5, 50]$, following the practice of Top-$K$ shifted by $N$ \cite{moreira2025nvretrieverimprovingtextembedding}. We set $\lambda_{1}$, $\lambda_{2}$ and $\lambda_{3}$ as $0.5$, $0.2$ and $0.3$ respectively.

\subsubsection{Evaluation}
\label{sec:evaluation}
We evaluate our approach on retrieval tasks in a zero-shot setting on the test splits of MSR-VTT, DiDeMo, and MSVD. Following standard protocol, we use one caption per video and report Recall@$1$, $5$, and $10$ for both text-to-video and video-to-text retrieval, except for MSVD where the standard protocol for video-to-text retrieval accounts for multiple possible captions per video. When reporting results with re-ranking, we first retrieve the top-50 candidates for each test query using \textit{VeRVE-Embed} through cosine similarity search. We then apply \textit{VeRVE-Ranker} to these candidates and re-rank them according to the predicted confidence scores to produce the final ranking. When reporting embedding-only numbers, we rank candidates directly using the cosine similarities from \textit{VeRVE-Embed} and compute metrics on these ranks.

\subsubsection{Results}

\begingroup
\renewcommand{\thefootnote}{\fnsymbol{footnote}}
\footnotetext[2]{This denotes the embedding-only performance of InternVideo2 with its vision-text matching head disabled as reported in \cite{guo2025universalvideoretrievalgeneralizing}.}
\endgroup

\begin{table}[t]
\centering
\caption{Zero-shot Retrieval on MSR-VTT 1K test set. \textbf{Bold} indicates the best and \underline{underline} the second-best result under each setting. \textit{VeRVE} demonstrates strong performance under both settings.}
\label{tab:msrvtt}
\small
\scalebox{0.94}{
\begin{tabular}{l|cc|cc}
\toprule
\multirow{2}{*}{\textbf{Model}} & \multicolumn{2}{c|}{\textbf{T→V}} & \multicolumn{2}{c}{\textbf{V→T}} \\
\cmidrule(lr){2-3} \cmidrule(lr){4-5}
& \textbf{R@1} & \textbf{R@5} & \textbf{R@1} & \textbf{R@5} \\
\midrule
\rowcolor{gray!20}
\multicolumn{5}{l}{\textit{Embedding-only models}} \\
Long-CLIP L/14 \cite{zhang2024longclipunlockinglongtextcapability} & 40.9 & 65.5 & 36.2 & 62.2 \\
LanguageBind \cite{zhu2024languagebindextendingvideolanguagepretraining} & 42.1 & 65.9 & 40.1 & 65.4 \\
InternVideo2-Embed 6B\textsuperscript{$\dagger$} \cite{wang2024internvideo2scalingfoundationmodels} & \textbf{48.5} & -- & -- & -- \\
VLM2Vec 7B \cite{jiang2025vlm2vectrainingvisionlanguagemodels} & 43.5 & -- & -- & -- \\
VLM2Vec-V2 2B \cite{meng2025vlm2vecv2advancingmultimodalembedding} & 33.0 & -- & -- & -- \\
GVE-7B \cite{guo2025universalvideoretrievalgeneralizing} & 46.4 & -- & -- & -- \\
CaRe 7B \cite{xu2025carebenchfinegrainedbenchmarkvideo} & {43.9} & {67.0} & {41.7} & {68.1} \\
\textit{VeRVE-Embed 7B (Ours) }& \underline{46.8} & {70.8} & {42.6} & {65.8} \\
\midrule
\rowcolor{gray!20}
\multicolumn{5}{l}{\textit{Re-ranker models}} \\
LamRA 7B \cite{liu2024lamralargemultimodalmodel} & 44.7 & 68.6 & -- & -- \\
InternVideo2 6B \cite{wang2024internvideo2scalingfoundationmodels} & \textbf{55.9} & \textbf{78.3} & \textbf{53.7} & \textbf{77.5} \\
\textit{VeRVE 7B (Ours)} & \underline{52.4} & \underline{76.0} & \underline{47.0} & \underline{69.9} \\
\bottomrule
\end{tabular}
}
\end{table}
\begin{table}[t]
\centering
\caption{Zero-shot Retrieval on DiDeMo test set. \textit{VeRVE} provides SOTA T→V retrieval, while \textit{VeRVE-Embed} is the best among embedding-only models.}
\label{tab:didemo}
\small
\scalebox{0.94}{
\begin{tabular}{l|cc|cc}
\toprule
\multirow{2}{*}{\textbf{Model}} & \multicolumn{2}{c|}{\textbf{T→V}} & \multicolumn{2}{c}{\textbf{V→T}} \\
\cmidrule(lr){2-3} \cmidrule(lr){4-5}
& \textbf{R@1} & \textbf{R@5} & \textbf{R@1} & \textbf{R@5} \\
\midrule
\rowcolor{gray!20}
\multicolumn{5}{l}{\textit{Embedding-only models}} \\
Long-CLIP L/14 \cite{zhang2024longclipunlockinglongtextcapability} & 32.4 & 56.2 & 28.5 & 54.1 \\
LanguageBind \cite{zhu2024languagebindextendingvideolanguagepretraining} & 35.6 & 63.6 & 35.6 & 62.8 \\
InternVideo2-Embed 6B\textsuperscript{$\dagger$} \cite{wang2024internvideo2scalingfoundationmodels} & 41.8 & -- & -- & -- \\
VLM2Vec-V2 2B \cite{meng2025vlm2vecv2advancingmultimodalembedding} & 29.9 & -- & -- & -- \\
GVE-7B \cite{guo2025universalvideoretrievalgeneralizing} & \underline{43.3} & -- & -- & -- \\
CaRe 7B \cite{xu2025carebenchfinegrainedbenchmarkvideo} & {41.4} & {68.5} & {39.1} & {66.0} \\
\textit{VeRVE-Embed 7B (Ours)} & \textbf{46.6} & 70.8 & 42.2 & 68.5 \\
\midrule
\rowcolor{gray!20}
\multicolumn{5}{l}{\textit{Re-ranker models}} \\
InternVideo2 6B \cite{wang2024internvideo2scalingfoundationmodels} & \underline{57.9} & \underline{80.0} & \textbf{57.1} & \textbf{79.9} \\
\textit{VeRVE 7B (Ours)} & \textbf{58.8} & \textbf{81.0} & \underline{52.5} & \underline{79.0} \\
\bottomrule
\end{tabular}
}
\end{table}

We evaluate our approach on zero-shot video-text retrieval across three benchmarks: MSR-VTT~\cite{7780940}, DiDeMo~\cite{hendricks2017localizingmomentsvideonatural}, and MSVD~\cite{chen-dolan-2011-collecting}. Results are presented in \cref{tab:msrvtt,tab:didemo,tab:msvd} for both text-to-video (T2V) and video-to-text (V2T) retrieval. We compare models under two configurations: \textit{embedding-only} models that perform retrieval solely through embedding-based similarity search, and \textit{joint-processing} models that employ re-ranking or joint query-candidate processing to assign matching scores. 

\begin{table}[t]
\centering
\caption{Zero-shot Retrieval on MSVD test set.}
\label{tab:msvd}
\small
\scalebox{0.94}{
\begin{tabular}{l|cc|cc}
\toprule
\multirow{2}{*}{\textbf{Model}} & \multicolumn{2}{c|}{\textbf{T→V}} & \multicolumn{2}{c}{\textbf{V→T}} \\
\cmidrule(lr){2-3} \cmidrule(lr){4-5}
& \textbf{R@1} & \textbf{R@5} & \textbf{R@1} & \textbf{R@5} \\
\midrule
\rowcolor{gray!20}
\multicolumn{5}{l}{\textit{Embedding-only models}} \\
Long-CLIP L/14 \cite{zhang2024longclipunlockinglongtextcapability} & 46.5 & 73.5 & 69.3 & 86.0 \\
LanguageBind \cite{zhu2024languagebindextendingvideolanguagepretraining} & \textbf{50.0} & 77.7 & 75.1 & 90 \\
VLM2Vec 7B \cite{jiang2025vlm2vectrainingvisionlanguagemodels} & 49.5 & -- & -- & -- \\
\textit{VeRVE-Embed 7B (Ours)} & \underline{49.8} & 77.6 & 73.4 & 87.0 \\
\midrule
\rowcolor{gray!20}
\multicolumn{5}{l}{\textit{Re-ranker models}} \\
LamRA 7B \cite{liu2024lamralargemultimodalmodel} & 52.4 & 79.8 & -- & -- \\
InternVideo2 6B \cite{wang2024internvideo2scalingfoundationmodels} & \textbf{59.3} & \textbf{84.4} & \textbf{83.1} & \textbf{94.2} \\
\textit{VeRVE 7B (Ours)} & \underline{57.8} & \underline{82.2} & \underline{79.0} & \underline{90.1} \\
\bottomrule
\end{tabular}
}
\end{table}
\begin{table}[t]
\centering
\caption{Zero-Shot Text + Video → Video Retrieval on CoVR test set. \textit{VeRVE-Embed} sets a new state of the art across all metrics.}
\label{tab:covr}
\vspace{-0.3cm}
\scalebox{0.9}{ 
\begin{tabular}{lccc}
\toprule
\textbf{Model Name} & \textbf{R@1} & \textbf{R@5} & \textbf{R@10} \\
\midrule
COVR-BLIP \cite{Ventura_2024} & 45.46 & 70.46 & 79.54 \\
Thawakar et. al \cite{thawakar2024composedvideoretrievalenriched} & 47.52 & 72.18 & 82.37 \\
\textit{VeRVE-Embed 7B (Ours)} & \textbf{55.49} & \textbf{79.41} & \textbf{86.26} \\
\bottomrule
\end{tabular}}
\end{table}

\textbf{Embedding-only performance.} Our \textit{VeRVE-Embed} model, which performs retrieval through embedding-based lookup, consistently outperforms other MLLM-based approaches including VLM2Vec~\cite{jiang2025vlm2vectrainingvisionlanguagemodels}, CaRe~\cite{xu2025carebenchfinegrainedbenchmarkvideo} and GVE \cite{guo2025universalvideoretrievalgeneralizing}, across all benchmarks. On DiDeMo, \textit{VeRVE-Embed} achieves 46.6\% R@1 for T2V retrieval, even outperforming InternVideo2 by 4.8 percentage points. 

\textbf{Joint-processing performance.} We evaluate our dual-stage system termed \textit{VeRVE} which retrieves top-$K$ candidates using \textit{VeRVE-Embed} and refines them using \textit{VeRVE-Ranker}. Our approach substantially outperforms LamRA~\cite{liu2024lamralargemultimodalmodel}, another MLLM-based dual-stage method on MSR-VTT with achieving +7.7 points over LamRA (T2V R@1), and +5.4 points on MSVD. 
On DiDeMo, our model outperforms InternVideo2 by 0.9 points on T2V R@1, despite InternVideo2 being a specialized SOTA video retrieval model trained on $\sim400$M multimodal samples spanning images, videos, text, and audio and using pairwise joint query-candidate processing with a matching head for scoring.
For V2T retrieval, our method trails InternVideo2 by 4-7 points on R@1 across benchmarks, though it outperforms all single-stage approaches. Critically, we achieve this competitive performance using only ${\sim}700$K adaptation samples, out of which only ${\sim}105$K samples are videos.

\textbf{Inference computational costs.} Within the embedding-only paradigm, our approach incurs no additional processing cost compared to MLLM-based methods such as VLM2Vec, CaRe and GVE. Specifically, we use the same-sized 7B LLM backbone requiring only one forward pass to obtain the same-sized $3584$ dimensional embedding. Within the ranker-based paradigm, our method re-ranks the top-$50$ candidates, consistent with LamRA, needing $50\times m$ comparisons for $m$ queries and therefore does not increase retrieval-time complexity. By contrast, InternVideo2 applies joint video-text processing with a matching head to all 
$m \times n$ query-target pairs in the corpus. As a result, while InternVideo2 achieves strong retrieval performance, it does so at a high inference cost.
\begin{table*}[t]
\centering
\caption{Zero-shot video moment localization results on Charades-STA and ActivityNet Captions. ``Train Samples" indicates the number of temporal annotation pairs (time intervals and language queries) used during training. \textdagger \; trained on raw videos from these datasets albeit without annotations. Our approach represents pure zero-shot performance requiring no temporal annotations during training.}
\label{tab:moment}
\small
\scalebox{0.92}{
\begin{tabular}{lc|cccc|cccc}
\toprule
\textbf{Method} & \textbf{\# Train}  & \multicolumn{4}{c|}{\textbf{Charades-STA}} & \multicolumn{4}{c}{\textbf{ActivityNet-Captions}} \\
& \textbf{Samples} & \textbf{R@0.3} & \textbf{R@0.5} & \textbf{R@0.7} & \textbf{mIoU} & \textbf{R@0.3} & \textbf{R@0.5} & \textbf{R@0.7} & \textbf{mIoU} \\
\midrule
\rowcolor{gray!20}
\multicolumn{10}{l}{\textit{Specialist models}} \\
UniVTG ~\cite{lin2023univtgunifiedvideolanguagetemporal} & 4.2M  & 44.1 & 25.2 & 10.0 & 27.1 & -- & -- & -- & -- \\
SeViLA ~\cite{sevila} & 129M  & -- & -- & -- & -- & 31.6 & 19.0 & 10.1 & 23.0 \\
LT-ZVG \textdagger ~\cite{kim2022languagefreetrainingzeroshotvideo} & --  & \underline{52.9} & \textbf{37.2} & \textbf{19.3} & \textbf{36.0} & \underline{47.6} & \textbf{32.6} & \textbf{15.4} & 31.8 \\
\midrule
\rowcolor{gray!20}
\multicolumn{10}{l}{\textit{MLLM-based models}} \\
VideoChat2 ~\cite{li2024mvbenchcomprehensivemultimodalvideo} & 2M  & 38.0 & 14.3 & 3.8 & 24.6 & 40.8 & 27.8 & 9.3 & 27.9 \\
Momenter ~\cite{qian2024momentoradvancingvideolarge} & 10M  & 42.6 & 26.6 & 11.6 & 28.5 & 42.9 & 23.0 & 12.4 & 29.3 \\
TimeLLM~\cite{huang2023vtimellmempowerllmgrasp} & 170K  & 51.0 & 27.5 & 11.4 & 31.2 & 44.0 & 27.8 & 14.3 & 30.4 \\
TimeChat ~\cite{ren2024timechattimesensitivemultimodallarge} & 125K  & -- & 32.2 & 13.4 & -- & -- & -- & -- & -- \\
HawkEye ~\cite{wang2024hawkeyetrainingvideotextllms} & 715K  & 50.6 & 31.4 & 14.5 & 33.7 & \textbf{49.1} & \underline{29.3} & 10.7 & \underline{32.7} \\
ChatVTG ~\cite{qu2024chatvtgvideotemporalgrounding} & 100K  & {52.7} & {33.0} & {15.9} & {34.9} & 40.7 & 22.5 & 9.4 & 27.2 \\
\midrule
\textit{VeRVE-Embed 7B (Ours)}  & --  & \textbf{55.5} & \underline{36.8} & \underline{16.6} & \underline{35.9} & {47.2} & {26.7} & \underline{14.5} & \textbf{33.4} \\
\bottomrule
\end{tabular}
}
\end{table*}

\begin{table}[t]
\centering
\caption{Ablation Studies on MSR-VTT 1K-A test set for Text-to-Video Retrieval}
\label{tab:ablation}
\vspace{-0.3cm}
\scalebox{0.9}{
\begin{tabular}{lcc}
\toprule
\textbf{Configuration} & \textbf{R@1} & \textbf{R@5} \\
\midrule
\multicolumn{3}{l}{\textit{Contrastive training data}} \\
Image only (CC595k) & 44.0 & 69.4 \\
Video only (PEVideo) & 41.6 & 64.7 \\
\textit{VeRVE-Embed} & 46.8 & 70.8 \\
\midrule
\multicolumn{3}{l}{\textit{Socratic baseline}} \\
Caption-based Retrieval & 32.6 & 54.9 \\
\midrule
\multicolumn{3}{l}{\textit{Re-ranking Loss Components}} \\
Random negatives only & 55.3 & 76.6 \\
Hard negatives only & 52.2 & 73.6 \\
Preference loss only & 53.1 & 75.5 \\
\bottomrule
\end{tabular}}
\vspace{-0.3cm}
\end{table}

\subsection{Composed Video Retrieval}

We evaluate our corpus-level embedding model on the CoVR benchmark \cite{Ventura_2024} for composed video retrieval in a zero-shot setting. CoVR consists of triplets in the form (source video, modification text, target video), where the task is to retrieve the target video representing the transformation the source video would undergo under the textual modification. This task is challenging as it requires jointly interpreting the visual content of the source video and the semantic transformation described in the modification text.
As shown in \cref{tab:covr}, our approach achieves state-of-the-art performance, outperforming Thawakar et. al \cite{thawakar2024composedvideoretrievalenriched} by $\sim$8\% on R@1, highlighting the effectiveness of our zero-shot composed retrieval formulation.

\subsection{Zero-shot Moment Retrieval}

We evaluate our embedding model on moment retrieval using Charades-STA~\cite{gao2017talltemporalactivitylocalization} and ActivityNet-Captions~\cite{krishna2017densecaptioningeventsvideos}. Our approach, detailed in \cref{S:MOMENT}, operates by computing dense frame-level similarities and localizing high-confidence temporal regions through peak detection. Following standard convention, we report Recall@$k$ at IoU thresholds and mIoU metrics on the test and val-2 splits respectively.

\cref{tab:moment} compares our approach against competitive methods including generalist MLLM-based models and task-specific specialists (architectures tailored to the task) in a zero-shot setting. Our \textit{VeRVE-Embed} achieves competitive performance across both benchmarks in a \emph{purely} zero-shot manner. We obtain the best R@0.3 on Charades-STA and the best mIoU on ActivityNet-Captions, while achieving strong results across other metrics. Critically, this represents true zero-shot learning, in contrast to prior methods that, despite being evaluated zero-shot on target datasets, still leverage temporally-annotated samples from auxiliary moment localization datasets during training \cite{wang2024hawkeyetrainingvideotextllms}.

Our frame-level embedding approach offers several advantages: (1)~it is computationally efficient compared to MLLM-based methods that generate frame captions or jointly process entire videos to predict timestamps, enabling better scalability to long-form videos; (2)~it achieves strong performance without auxiliary post-processing steps such as query debiasing with external LLMs or candidate proposal generation from separate models, which are commonly used to boost performance in prior work; (3)~We predict variable-sized windows in a single step in contrast to prior work \cite{qu2024chatvtgvideotemporalgrounding} that undergoes multi-step window refinements; and (4)~since our model naturally handles composed queries, it can be readily extended to composed moment retrieval tasks, which we leave for future exploration.
\subsection{Ablations}
We conduct a series of ablation studies on the MSR-VTT benchmark in a zero-shot setting to analyze the contribution of individual components in our framework. Results are summarized in \cref{tab:ablation}.

\textbf{Contrastive training data.} We ablate the design choice of our contrastive training process by evaluating models trained with only image-text data (CC595K) or only video-text data (PEVideo). The results demonstrate that both stages contribute complementary benefits: image-only training achieves 44.0\% R@1 while video-only training achieves 41.6\% R@1, compared to 46.8\% R@1 for our final \textit{VeRVE-Embed} model, yielding absolute improvements of 2.8 and 5.2 percentage points respectively. 

\textbf{Socratic baseline.} To assess the efficacy of our cross-modal contrastive training strategy, we implement a caption-based retrieval baseline that approximates raw zero-shot performance of our base MLLM. We generate detailed video captions using Qwen2.5-VL, embed captions and queries using GRIT-LM 7B~\cite{muennighoff2025generativerepresentationalinstructiontuning}, a SOTA text embedder, and compute retrieval metrics. This baseline achieves 32.6\% R@1, substantially lower than our embedding-only model (46.8\%), highlighting the value of cross-modal contrastive learning over caption-based retrieval.

\textbf{Re-ranking loss components.} We ablate the individual loss components in our re-ranker training. Training with random negatives achieves 55.3\% R@1, while hard negatives alone achieve 52.2\%, and preference loss alone achieves 53.1\%.

\section{Conclusion}
We introduced \textit{VeRVE}, a video retrieval-centric adaptation of an MLLM that unifies corpus-level retrieval, candidate re-ranking, composed multimodal querying and moment localization within a single architecture. Through contrastive alignment on video–text pairs, \textit{VeRVE} produces strong multimodal embeddings and further improves retrieval precision with a simple reranker. Despite relying on modest adaptation data consisting of vision-text pairs, \textit{VeRVE} matches or exceeds the performance of specialized video retrieval systems while enabling versatile retrieval modes such as composed and moment retrieval. Our results suggest that retrieval-oriented contrastive adaptation of MLLMs is a powerful and scalable design choice, and we anticipate that training on larger video–text corpora and richer video descriptions will further elevate the capability of unified retrieval systems. We hope \textit{VeRVE} serves as a foundation for future work developing flexible, general-purpose video retrieval engines that can handle increasingly complex and expressive user queries.
{
    \small
    \bibliographystyle{ieeenat_fullname}
    \bibliography{main}

@String(CVPR= {IEEE Conf. Comput. Vis. Pattern Recog.})

@String(CVPR  = {CVPR})

@misc{guo2025universalvideoretrievalgeneralizing,
      title={Towards Universal Video Retrieval: Generalizing Video Embedding via Synthesized Multimodal Pyramid Curriculum}, 
      author={Zhuoning Guo et. al.},
      year={2025},
      eprint={2510.27571},
      archivePrefix={arXiv},
      primaryClass={cs.CV},
      url={https://arxiv.org/abs/2510.27571}, 
}

@misc{meng2025vlm2vecv2advancingmultimodalembedding,
      title={VLM2Vec-V2: Advancing Multimodal Embedding for Videos, Images, and Visual Documents}, 
      author={Rui Meng et. al.},
      year={2025},
      eprint={2507.04590},
      archivePrefix={arXiv},
      primaryClass={cs.CV},
      url={https://arxiv.org/abs/2507.04590}, 
}

@article{Ventura_2024,
   title={CoVR-2: Automatic Data Construction for Composed Video Retrieval},
   volume={46},
   ISSN={1939-3539},
   url={http://dx.doi.org/10.1109/TPAMI.2024.3463799},
   DOI={10.1109/tpami.2024.3463799},
   number={12},
   journal={IEEE Transactions on Pattern Analysis and Machine Intelligence},
   publisher={Institute of Electrical and Electronics Engineers (IEEE)},
   author={Ventura, Lucas and Yang, Antoine and Schmid, Cordelia and Varol, Gül},
   year={2024},
   month=dec, pages={11409–11421} }

@misc{hummel2024egocvregocentricbenchmarkfinegrained,
      title={EgoCVR: An Egocentric Benchmark for Fine-Grained Composed Video Retrieval}, 
      author={Thomas Hummel and Shyamgopal Karthik and Mariana-Iuliana Georgescu and Zeynep Akata},
      year={2024},
      eprint={2407.16658},
      archivePrefix={arXiv},
      primaryClass={cs.CV},
      url={https://arxiv.org/abs/2407.16658}, 
}

@misc{moreira2025nvretrieverimprovingtextembedding,
      title={NV-Retriever: Improving text embedding models with effective hard-negative mining}, 
      author={Gabriel de Souza P. Moreira and Radek Osmulski and Mengyao Xu and Ronay Ak and Benedikt Schifferer and Even Oldridge},
      year={2025},
      eprint={2407.15831},
      archivePrefix={arXiv},
      primaryClass={cs.IR},
      url={https://arxiv.org/abs/2407.15831}, 
}

@misc{zhang2025llavavideovideoinstructiontuning,
      title={LLaVA-Video: Video Instruction Tuning With Synthetic Data}, 
      author={Yuanhan Zhang and Jinming Wu and Wei Li and Bo Li and Zejun Ma and Ziwei Liu and Chunyuan Li},
      year={2025},
      eprint={2410.02713},
      archivePrefix={arXiv},
      primaryClass={cs.CV},
      url={https://arxiv.org/abs/2410.02713}, 
}

@InProceedings{li2022blipbootstrappinglanguageimagepretraining,
  title = 	 {{BLIP}: Bootstrapping Language-Image Pre-training for Unified Vision-Language Understanding and Generation},
  author =       {Li, Junnan and Li, Dongxu and Xiong, Caiming and Hoi, Steven},
  booktitle = 	 {Proceedings of the 39th International Conference on Machine Learning},
  pages = 	 {12888--12900},
  year = 	 {2022},
  editor = 	 {Chaudhuri, Kamalika and Jegelka, Stefanie and Song, Le and Szepesvari, Csaba and Niu, Gang and Sabato, Sivan},
  volume = 	 {162},
  series = 	 {Proceedings of Machine Learning Research},
  month = 	 {17--23 Jul},
  publisher =    {PMLR},
  url = 	 {https://proceedings.mlr.press/v162/li22n.html}
}

@misc{li2023blip2bootstrappinglanguageimagepretraining,
      title={BLIP-2: Bootstrapping Language-Image Pre-training with Frozen Image Encoders and Large Language Models}, 
      author={Junnan Li and Dongxu Li and Silvio Savarese and Steven Hoi},
      year={2023},
      eprint={2301.12597},
      archivePrefix={arXiv},
      primaryClass={cs.CV},
      url={https://arxiv.org/abs/2301.12597}, 
}

@misc{zhang2025gmeimprovinguniversalmultimodal,
      title={GME: Improving Universal Multimodal Retrieval by Multimodal LLMs}, 
      author={Xin Zhang and Yanzhao Zhang and Wen Xie and Mingxin Li and Ziqi Dai and Dingkun Long and Pengjun Xie and Meishan Zhang and Wenjie Li and Min Zhang},
      year={2025},
      eprint={2412.16855},
      archivePrefix={arXiv},
      primaryClass={cs.CL},
      url={https://arxiv.org/abs/2412.16855}, 
}

@misc{liu2024lamralargemultimodalmodel,
      title={LamRA: Large Multimodal Model as Your Advanced Retrieval Assistant}, 
      author={Yikun Liu and Pingan Chen and Jiayin Cai and Xiaolong Jiang and Yao Hu and Jiangchao Yao and Yanfeng Wang and Weidi Xie},
      year={2024},
      eprint={2412.01720},
      archivePrefix={arXiv},
      primaryClass={cs.CV},
      url={https://arxiv.org/abs/2412.01720}, 
}

@misc{wang2024internvideo2scalingfoundationmodels,
      title={InternVideo2: Scaling Foundation Models for Multimodal Video Understanding}, 
      author={Yi Wang and Kunchang Li and Xinhao Li and Jiashuo Yu and Yinan He and Chenting Wang and Guo Chen and Baoqi Pei and Ziang Yan and Rongkun Zheng and Jilan Xu and Zun Wang and Yansong Shi and Tianxiang Jiang and Songze Li and Hongjie Zhang and Yifei Huang and Yu Qiao and Yali Wang and Limin Wang},
      year={2024},
      eprint={2403.15377},
      archivePrefix={arXiv},
      primaryClass={cs.CV},
      url={https://arxiv.org/abs/2403.15377}, 
}

@INPROCEEDINGS{7780940,
  author={Xu, Jun and Mei, Tao and Yao, Ting and Rui, Yong},
  booktitle={2016 IEEE Conference on Computer Vision and Pattern Recognition (CVPR)}, 
  title={MSR-VTT: A Large Video Description Dataset for Bridging Video and Language}, 
  year={2016},
  volume={},
  number={},
  pages={5288-5296},
  doi={10.1109/CVPR.2016.571}}

@inproceedings{chen-dolan-2011-collecting,
    title = "Collecting Highly Parallel Data for Paraphrase Evaluation",
    author = "Chen, David  and
      Dolan, William",
    editor = "Lin, Dekang  and
      Matsumoto, Yuji  and
      Mihalcea, Rada",
    booktitle = "Proceedings of the 49th Annual Meeting of the Association for Computational Linguistics: Human Language Technologies",
    month = jun,
    year = "2011",
    address = "Portland, Oregon, USA",
    publisher = "Association for Computational Linguistics",
    url = "https://aclanthology.org/P11-1020/",
    pages = "190--200"
}

@misc{hendricks2017localizingmomentsvideonatural,
      title={Localizing Moments in Video with Natural Language}, 
      author={Lisa Anne Hendricks and Oliver Wang and Eli Shechtman and Josef Sivic and Trevor Darrell and Bryan Russell},
      year={2017},
      eprint={1708.01641},
      archivePrefix={arXiv},
      primaryClass={cs.CV},
      url={https://arxiv.org/abs/1708.01641}, 
}

@misc{gao2017talltemporalactivitylocalization,
      title={TALL: Temporal Activity Localization via Language Query}, 
      author={Jiyang Gao and Chen Sun and Zhenheng Yang and Ram Nevatia},
      year={2017},
      eprint={1705.02101},
      archivePrefix={arXiv},
      primaryClass={cs.CV},
      url={https://arxiv.org/abs/1705.02101}, 
}

@misc{alayrac2022flamingovisuallanguagemodel,
      title={Flamingo: a Visual Language Model for Few-Shot Learning}, 
      author={Jean-Baptiste Alayrac and Jeff Donahue and Pauline Luc and Antoine Miech and Iain Barr and Yana Hasson and Karel Lenc and Arthur Mensch and Katie Millican and Malcolm Reynolds and Roman Ring and Eliza Rutherford and Serkan Cabi and Tengda Han and Zhitao Gong and Sina Samangooei and Marianne Monteiro and Jacob Menick and Sebastian Borgeaud and Andrew Brock and Aida Nematzadeh and Sahand Sharifzadeh and Mikolaj Binkowski and Ricardo Barreira and Oriol Vinyals and Andrew Zisserman and Karen Simonyan},
      year={2022},
      eprint={2204.14198},
      archivePrefix={arXiv},
      primaryClass={cs.CV},
      url={https://arxiv.org/abs/2204.14198}, 
}

@misc{liu2023visualinstructiontuning,
      title={Visual Instruction Tuning}, 
      author={Haotian Liu and Chunyuan Li and Qingyang Wu and Yong Jae Lee},
      year={2023},
      eprint={2304.08485},
      archivePrefix={arXiv},
      primaryClass={cs.CV},
      url={https://arxiv.org/abs/2304.08485}, 
}

@misc{wang2024cogvlmvisualexpertpretrained,
      title={CogVLM: Visual Expert for Pretrained Language Models}, 
      author={Weihan Wang and Qingsong Lv and Wenmeng Yu and Wenyi Hong and Ji Qi and Yan Wang and Junhui Ji and Zhuoyi Yang and Lei Zhao and Xixuan Song and Jiazheng Xu and Bin Xu and Juanzi Li and Yuxiao Dong and Ming Ding and Jie Tang},
      year={2024},
      eprint={2311.03079},
      archivePrefix={arXiv},
      primaryClass={cs.CV},
      url={https://arxiv.org/abs/2311.03079}, 
}

@inproceedings{chen2024internvl,
    title={Internvl: Scaling up vision foundation models and aligning for generic visual-linguistic tasks},
    author={Chen, Zhe and Wu, Jiannan and Wang, Wenhai and Su, Weijie and Chen, Guo and Xing, Sen and Zhong, Muyan and Zhang, Qinglong and Zhu, Xizhou and Lu, Lewei and others},
    booktitle={Proceedings of the IEEE/CVF Conference on Computer Vision and Pattern Recognition},
    pages={24185--24198},
    year={2024}
  }

@misc{bai2025qwen25vltechnicalreport,
      title={Qwen2.5-VL Technical Report}, 
      author={Shuai Bai and Keqin Chen and Xuejing Liu and Jialin Wang and Wenbin Ge and Sibo Song and Kai Dang and Peng Wang and Shijie Wang and Jun Tang and Humen Zhong and Yuanzhi Zhu and Mingkun Yang and Zhaohai Li and Jianqiang Wan and Pengfei Wang and Wei Ding and Zheren Fu and Yiheng Xu and Jiabo Ye and Xi Zhang and Tianbao Xie and Zesen Cheng and Hang Zhang and Zhibo Yang and Haiyang Xu and Junyang Lin},
      year={2025},
      eprint={2502.13923},
      archivePrefix={arXiv},
      primaryClass={cs.CV},
      url={https://arxiv.org/abs/2502.13923}, 
}

@misc{miech2019howto100mlearningtextvideoembedding,
      title={HowTo100M: Learning a Text-Video Embedding by Watching Hundred Million Narrated Video Clips}, 
      author={Antoine Miech and Dimitri Zhukov and Jean-Baptiste Alayrac and Makarand Tapaswi and Ivan Laptev and Josef Sivic},
      year={2019},
      eprint={1906.03327},
      archivePrefix={arXiv},
      primaryClass={cs.CV},
      url={https://arxiv.org/abs/1906.03327}, 
}

@misc{miech2020endtoendlearningvisualrepresentations,
      title={End-to-End Learning of Visual Representations from Uncurated Instructional Videos}, 
      author={Antoine Miech and Jean-Baptiste Alayrac and Lucas Smaira and Ivan Laptev and Josef Sivic and Andrew Zisserman},
      year={2020},
      eprint={1912.06430},
      archivePrefix={arXiv},
      primaryClass={cs.CV},
      url={https://arxiv.org/abs/1912.06430}, 
}

@misc{luo2021clip4clipempiricalstudyclip,
      title={CLIP4Clip: An Empirical Study of CLIP for End to End Video Clip Retrieval}, 
      author={Huaishao Luo and Lei Ji and Ming Zhong and Yang Chen and Wen Lei and Nan Duan and Tianrui Li},
      year={2021},
      eprint={2104.08860},
      archivePrefix={arXiv},
      primaryClass={cs.CV},
      url={https://arxiv.org/abs/2104.08860}, 
}

@misc{bain2022frozentimejointvideo,
      title={Frozen in Time: A Joint Video and Image Encoder for End-to-End Retrieval}, 
      author={Max Bain and Arsha Nagrani and Gül Varol and Andrew Zisserman},
      year={2022},
      eprint={2104.00650},
      archivePrefix={arXiv},
      primaryClass={cs.CV},
      url={https://arxiv.org/abs/2104.00650}, 
}

@misc{bain2022cliphitchhikersguidelongvideo,
      title={A CLIP-Hitchhiker's Guide to Long Video Retrieval}, 
      author={Max Bain and Arsha Nagrani and Gül Varol and Andrew Zisserman},
      year={2022},
      eprint={2205.08508},
      archivePrefix={arXiv},
      primaryClass={cs.CV},
      url={https://arxiv.org/abs/2205.08508}, 
}

@misc{rizve2024vidlavideolanguagealignmentscale,
      title={VidLA: Video-Language Alignment at Scale}, 
      author={Mamshad Nayeem Rizve and Fan Fei and Jayakrishnan Unnikrishnan and Son Tran and Benjamin Z. Yao and Belinda Zeng and Mubarak Shah and Trishul Chilimbi},
      year={2024},
      eprint={2403.14870},
      archivePrefix={arXiv},
      primaryClass={cs.CV},
      url={https://arxiv.org/abs/2403.14870}, 
}

@misc{cicchetti2025gramianmultimodalrepresentationlearning,
      title={Gramian Multimodal Representation Learning and Alignment}, 
      author={Giordano Cicchetti and Eleonora Grassucci and Luigi Sigillo and Danilo Comminiello},
      year={2025},
      eprint={2412.11959},
      archivePrefix={arXiv},
      primaryClass={cs.CV},
      url={https://arxiv.org/abs/2412.11959}, 
}

@misc{jiang2024e5vuniversalembeddingsmultimodal,
      title={E5-V: Universal Embeddings with Multimodal Large Language Models}, 
      author={Ting Jiang and Minghui Song and Zihan Zhang and Haizhen Huang and Weiwei Deng and Feng Sun and Qi Zhang and Deqing Wang and Fuzhen Zhuang},
      year={2024},
      eprint={2407.12580},
      archivePrefix={arXiv},
      primaryClass={cs.CL},
      url={https://arxiv.org/abs/2407.12580}, 
}

@misc{jiang2025vlm2vectrainingvisionlanguagemodels,
      title={VLM2Vec: Training Vision-Language Models for Massive Multimodal Embedding Tasks}, 
      author={Ziyan Jiang and Rui Meng and Xinyi Yang and Semih Yavuz and Yingbo Zhou and Wenhu Chen},
      year={2025},
      eprint={2410.05160},
      archivePrefix={arXiv},
      primaryClass={cs.CV},
      url={https://arxiv.org/abs/2410.05160}, 
}

@misc{xu2025carebenchfinegrainedbenchmarkvideo,
      title={CaReBench: A Fine-Grained Benchmark for Video Captioning and Retrieval}, 
      author={Yifan Xu and Xinhao Li and Yichun Yang and Desen Meng and Rui Huang and Limin Wang},
      year={2025},
      eprint={2501.00513},
      archivePrefix={arXiv},
      primaryClass={cs.CV},
      url={https://arxiv.org/abs/2501.00513}, 
}

@misc{radford2021learningtransferablevisualmodels,
      title={Learning Transferable Visual Models From Natural Language Supervision}, 
      author={Alec Radford and Jong Wook Kim and Chris Hallacy and Aditya Ramesh and Gabriel Goh and Sandhini Agarwal and Girish Sastry and Amanda Askell and Pamela Mishkin and Jack Clark and Gretchen Krueger and Ilya Sutskever},
      year={2021},
      eprint={2103.00020},
      archivePrefix={arXiv},
      primaryClass={cs.CV},
      url={https://arxiv.org/abs/2103.00020}, 
}

@misc{hu2021loralowrankadaptationlarge,
      title={LoRA: Low-Rank Adaptation of Large Language Models}, 
      author={Edward J. Hu and Yelong Shen and Phillip Wallis and Zeyuan Allen-Zhu and Yuanzhi Li and Shean Wang and Lu Wang and Weizhu Chen},
      year={2021},
      eprint={2106.09685},
      archivePrefix={arXiv},
      primaryClass={cs.CL},
      url={https://arxiv.org/abs/2106.09685}, 
}

@misc{oord2019representationlearningcontrastivepredictive,
      title={Representation Learning with Contrastive Predictive Coding}, 
      author={Aaron van den Oord and Yazhe Li and Oriol Vinyals},
      year={2019},
      eprint={1807.03748},
      archivePrefix={arXiv},
      primaryClass={cs.LG},
      url={https://arxiv.org/abs/1807.03748}, 
}

@misc{bolya2025perceptionencoderbestvisual,
      title={Perception Encoder: The best visual embeddings are not at the output of the network}, 
      author={Daniel Bolya and Po-Yao Huang and Peize Sun and Jang Hyun Cho and Andrea Madotto and Chen Wei and Tengyu Ma and Jiale Zhi and Jathushan Rajasegaran and Hanoona Rasheed and Junke Wang and Marco Monteiro and Hu Xu and Shiyu Dong and Nikhila Ravi and Daniel Li and Piotr Dollár and Christoph Feichtenhofer},
      year={2025},
      eprint={2504.13181},
      archivePrefix={arXiv},
      primaryClass={cs.CV},
      url={https://arxiv.org/abs/2504.13181}, 
}

@article{bradley1952RankAnalysisofIncompleteBlockDesigns,
 ISSN = {00063444, 14643510},
 URL = {http://www.jstor.org/stable/2334029},
 author = {Ralph Allan Bradley and Milton E. Terry},
 journal = {Biometrika},
 number = {3/4},
 pages = {324--345},
 publisher = {[Oxford University Press, Biometrika Trust]},
 title = {Rank Analysis of Incomplete Block Designs: I. The Method of Paired Comparisons},
 urldate = {2025-11-03},
 volume = {39},
 year = {1952}
}

@inproceedings{sharma-etal-2018-conceptual,
    title = "Conceptual Captions: A Cleaned, Hypernymed, Image Alt-text Dataset For Automatic Image Captioning",
    author = "Sharma, Piyush  and
      Ding, Nan  and
      Goodman, Sebastian  and
      Soricut, Radu",
    editor = "Gurevych, Iryna  and
      Miyao, Yusuke",
    booktitle = "Proceedings of the 56th Annual Meeting of the Association for Computational Linguistics (Volume 1: Long Papers)",
    month = jul,
    year = "2018",
    address = "Melbourne, Australia",
    publisher = "Association for Computational Linguistics",
    url = "https://aclanthology.org/P18-1238/",
    doi = "10.18653/v1/P18-1238",
    pages = "2556--2565"
}

@misc{muennighoff2025generativerepresentationalinstructiontuning,
      title={Generative Representational Instruction Tuning}, 
      author={Niklas Muennighoff and Hongjin Su and Liang Wang and Nan Yang and Furu Wei and Tao Yu and Amanpreet Singh and Douwe Kiela},
      year={2025},
      eprint={2402.09906},
      archivePrefix={arXiv},
      primaryClass={cs.CL},
      url={https://arxiv.org/abs/2402.09906}, 
}

@misc{lu2024llavamrlargelanguageandvisionassistant,
      title={LLaVA-MR: Large Language-and-Vision Assistant for Video Moment Retrieval}, 
      author={Weiheng Lu and Jian Li and An Yu and Ming-Ching Chang and Shengpeng Ji and Min Xia},
      year={2024},
      eprint={2411.14505},
      archivePrefix={arXiv},
      primaryClass={cs.CV},
      url={https://arxiv.org/abs/2411.14505}, 
}

@article{Meinardus2024TheSurprisingEffectivenessofMultimodal,
  title={The Surprising Effectiveness of Multimodal Large Language Models for Video Moment Retrieval},
  author={Boris Meinardus and Anil Batra and Anna Rohrbach and Marcus Rohrbach},
  journal={ArXiv},
  year={2024},
  volume={abs/2406.18113},
  url={https://api.semanticscholar.org/CorpusID:279618442}
}

@misc{ren2024timechattimesensitivemultimodallarge,
      title={TimeChat: A Time-sensitive Multimodal Large Language Model for Long Video Understanding}, 
      author={Shuhuai Ren and Linli Yao and Shicheng Li and Xu Sun and Lu Hou},
      year={2024},
      eprint={2312.02051},
      archivePrefix={arXiv},
      primaryClass={cs.CV},
      url={https://arxiv.org/abs/2312.02051}, 
}

@misc{qu2024chatvtgvideotemporalgrounding,
      title={ChatVTG: Video Temporal Grounding via Chat with Video Dialogue Large Language Models}, 
      author={Mengxue Qu and Xiaodong Chen and Wu Liu and Alicia Li and Yao Zhao},
      year={2024},
      eprint={2410.12813},
      archivePrefix={arXiv},
      primaryClass={cs.MM},
      url={https://arxiv.org/abs/2410.12813}, 
}

@misc{thawakar2024composedvideoretrievalenriched,
      title={Composed Video Retrieval via Enriched Context and Discriminative Embeddings}, 
      author={Omkar Thawakar and Muzammal Naseer and Rao Muhammad Anwer and Salman Khan and Michael Felsberg and Mubarak Shah and Fahad Shahbaz Khan},
      year={2024},
      eprint={2403.16997},
      archivePrefix={arXiv},
      primaryClass={cs.CV},
      url={https://arxiv.org/abs/2403.16997}, 
}

@misc{zhu2024languagebindextendingvideolanguagepretraining,
      title={LanguageBind: Extending Video-Language Pretraining to N-modality by Language-based Semantic Alignment}, 
      author={Bin Zhu and Bin Lin and Munan Ning and Yang Yan and Jiaxi Cui and HongFa Wang and Yatian Pang and Wenhao Jiang and Junwu Zhang and Zongwei Li and Wancai Zhang and Zhifeng Li and Wei Liu and Li Yuan},
      year={2024},
      eprint={2310.01852},
      archivePrefix={arXiv},
      primaryClass={cs.CV},
      url={https://arxiv.org/abs/2310.01852}, 
}

@misc{xu2024pllavaparameterfreellava,
      title={PLLaVA : Parameter-free LLaVA Extension from Images to Videos for Video Dense Captioning}, 
      author={Lin Xu and Yilin Zhao and Daquan Zhou and Zhijie Lin and See Kiong Ng and Jiashi Feng},
      year={2024},
      eprint={2404.16994},
      archivePrefix={arXiv},
      primaryClass={cs.CV},
      url={https://arxiv.org/abs/2404.16994}, 
}

@misc{shen2024longvuspatiotemporaladaptivecompression,
      title={LongVU: Spatiotemporal Adaptive Compression for Long Video-Language Understanding}, 
      author={Xiaoqian Shen and Yunyang Xiong and Changsheng Zhao and Lemeng Wu and Jun Chen and Chenchen Zhu and Zechun Liu and Fanyi Xiao and Balakrishnan Varadarajan and Florian Bordes and Zhuang Liu and Hu Xu and Hyunwoo J. Kim and Bilge Soran and Raghuraman Krishnamoorthi and Mohamed Elhoseiny and Vikas Chandra},
      year={2024},
      eprint={2410.17434},
      archivePrefix={arXiv},
      primaryClass={cs.CV},
      url={https://arxiv.org/abs/2410.17434}, 
}

@misc{krishna2017densecaptioningeventsvideos,
      title={Dense-Captioning Events in Videos}, 
      author={Ranjay Krishna and Kenji Hata and Frederic Ren and Li Fei-Fei and Juan Carlos Niebles},
      year={2017},
      eprint={1705.00754},
      archivePrefix={arXiv},
      primaryClass={cs.CV},
      url={https://arxiv.org/abs/1705.00754}, 
}

@misc{zhang2024longclipunlockinglongtextcapability,
      title={Long-CLIP: Unlocking the Long-Text Capability of CLIP}, 
      author={Beichen Zhang and Pan Zhang and Xiaoyi Dong and Yuhang Zang and Jiaqi Wang},
      year={2024},
      eprint={2403.15378},
      archivePrefix={arXiv},
      primaryClass={cs.CV},
      url={https://arxiv.org/abs/2403.15378}, 
}

@misc{qian2024momentoradvancingvideolarge,
      title={Momentor: Advancing Video Large Language Model with Fine-Grained Temporal Reasoning}, 
      author={Long Qian and Juncheng Li and Yu Wu and Yaobo Ye and Hao Fei and Tat-Seng Chua and Yueting Zhuang and Siliang Tang},
      year={2024},
      eprint={2402.11435},
      archivePrefix={arXiv},
      primaryClass={cs.CV},
      url={https://arxiv.org/abs/2402.11435}, 
}

@misc{lin2023univtgunifiedvideolanguagetemporal,
      title={UniVTG: Towards Unified Video-Language Temporal Grounding}, 
      author={Kevin Qinghong Lin and Pengchuan Zhang and Joya Chen and Shraman Pramanick and Difei Gao and Alex Jinpeng Wang and Rui Yan and Mike Zheng Shou},
      year={2023},
      eprint={2307.16715},
      archivePrefix={arXiv},
      primaryClass={cs.CV},
      url={https://arxiv.org/abs/2307.16715}, 
}

@misc{sevila,
      title={Self-Chained Image-Language Model for Video Localization and Question Answering}, 
      author={Shoubin Yu and Jaemin Cho and Prateek Yadav and Mohit Bansal},
      year={2023},
      eprint={2305.06988},
      archivePrefix={arXiv},
      primaryClass={cs.CV},
      url={https://arxiv.org/abs/2305.06988}, 
}

@misc{kim2022languagefreetrainingzeroshotvideo,
      title={Language-free Training for Zero-shot Video Grounding}, 
      author={Dahye Kim and Jungin Park and Jiyoung Lee and Seongheon Park and Kwanghoon Sohn},
      year={2022},
      eprint={2210.12977},
      archivePrefix={arXiv},
      primaryClass={cs.CV},
      url={https://arxiv.org/abs/2210.12977}, 
}

@misc{li2024mvbenchcomprehensivemultimodalvideo,
      title={MVBench: A Comprehensive Multi-modal Video Understanding Benchmark}, 
      author={Kunchang Li and Yali Wang and Yinan He and Yizhuo Li and Yi Wang and Yi Liu and Zun Wang and Jilan Xu and Guo Chen and Ping Luo and Limin Wang and Yu Qiao},
      year={2024},
      eprint={2311.17005},
      archivePrefix={arXiv},
      primaryClass={cs.CV},
      url={https://arxiv.org/abs/2311.17005}, 
}

@misc{huang2023vtimellmempowerllmgrasp,
      title={VTimeLLM: Empower LLM to Grasp Video Moments}, 
      author={Bin Huang and Xin Wang and Hong Chen and Zihan Song and Wenwu Zhu},
      year={2023},
      eprint={2311.18445},
      archivePrefix={arXiv},
      primaryClass={cs.CV},
      url={https://arxiv.org/abs/2311.18445}, 
}

@misc{wang2024hawkeyetrainingvideotextllms,
      title={im invincible im unstoppable i'm a lion}, 
      author={Yueqian Wang and Xiaojun Meng and Jianxin Liang and Yuxuan Wang and Qun Liu and Dongyan Zhao},
      year={2024},
      eprint={2403.10228},
      archivePrefix={arXiv},
      primaryClass={cs.CV},
      url={https://arxiv.org/abs/2403.10228}, 
}
}

\setcounter{figure}{0}
\setcounter{table}{0}
\setcounter{section}{0}

\renewcommand{\thetable}{\Alph{table}}
\renewcommand{\thefigure}{\Alph{figure}}
\renewcommand\thesection{\Alph{section}}
\clearpage
\makesupplementary

\section{Additional Implementation Details}
All VeRVE models use Qwen2.5-VL 7B \cite{bai2025qwen25vltechnicalreport} as the backbone architecture. We apply LoRA~\cite{hu2021loralowrankadaptationlarge} adapters to the query, key, value, and output projection layers within the LLM's self-attention modules, as well as to the MLP layers in the vision-language merger. We use a LoRA rank of 16 and scaling factor of 32.
For \textit{VeRVE-Embed} training, we use a learning rate of $2\mathrm{e}{-4}$
 for the image-text stage (stage
1) and $2\mathrm{e}{-5}$
 for the video-text stage (stage
2). We employ the AdamW optimizer with a cosine learning rate schedule and mixed-precision training (BF16). At inference time, we include dual-softmax based re-ordering before feeding the candidates to the re-ranker only for the \textit{VeRVE-Ranker} based results in \cref{tab:msrvtt,tab:didemo,tab:msvd}.

\subsection{Evaluation Datasets}

We provide detailed descriptions of all evaluation benchmarks used in our experiments. Dataset statistics are summarized in Table~\ref{tab:datasets}.

\subsubsection{Corpus-Level Video-Text Retrieval Datasets}

\textbf{MSR-VTT}~\cite{7780940} is a large-scale video description dataset containing video clips paired with natural language descriptions. The dataset covers diverse topics including human activities, sports, cooking, and entertainment. Following standard protocol, we evaluate on the 1K-A test split for zero-shot text-to-video and video-to-text retrieval, using one caption per video.

\noindent \textbf{DiDeMo}~\cite{hendricks2017localizingmomentsvideonatural} (Distinct Describable Moments) features 10,000 unedited videos, with each video containing multiple describable moments. Videos are annotated with natural language descriptions of specific temporal segments. The dataset is characterized by longer video durations (averaging 30 seconds) and paragraph-level captions that capture temporal progression. Following standard practice, we concatenate the sentences for each video and evaluate on the official test split for paragraph-to-video retrieval.

\noindent \textbf{MSVD}~\cite{chen-dolan-2011-collecting} (Microsoft Research Video Description) consists of 1,970 short video clips, each paired with approximately 40 human-annotated captions. The dataset focuses on single-action clips with clear visual content, making it a standard benchmark for evaluating video-language alignment. We follow the standard evaluation protocol, accounting for multiple captions per video in video-to-text retrieval metrics.

\subsubsection{Composed Video Retrieval Dataset}

\textbf{CoVR}~\cite{Ventura_2024} is a large-scale benchmark for composed video retrieval, consisting of automatically constructed triplets in the form (source video, modification text, target video). The dataset contains diverse modification types including object changes, scene transformations, action modifications, and style adjustments. Each triplet requires models to understand both the visual content of the source video and the semantic transformation described in the modification text. The test set contains high-quality manually verified examples. We evaluate using the standard metrics of Recall@1, 5, and 10 for the text+video$\rightarrow$video retrieval task.

\begin{table}[t]
\centering
\caption{Evaluation dataset statistics.}
\label{tab:datasets}
\small
\scalebox{0.9}{
\begin{tabular}{l|l|c|c}
\toprule
\textbf{Benchmark} & \textbf{Task} & \textbf{\#Queries} & \textbf{\#Videos}  \\
\midrule
MSR-VTT \cite{7780940} & T$\leftrightarrow$V & 1,000 & 1,000 \\
DiDeMo \cite{hendricks2017localizingmomentsvideonatural} & T$\leftrightarrow$V & 1,004 & 1,004 \\
MSVD \cite{chen-dolan-2011-collecting} & T$\leftrightarrow$V & 670 & 1,970 \\
CoVR-2 \cite{Ventura_2024} & TV$\rightarrow$V & 2,556 & 2,556 \\
Charades-STA \cite{gao2017talltemporalactivitylocalization} & T$\rightarrow$segment & 3,720 & 1,334 \\
ActivityNet-Captions \cite{krishna2017densecaptioningeventsvideos} & T$\rightarrow$segment & 17,031 & 4,885 \\
\bottomrule
\end{tabular}}
\end{table}

\subsubsection{Moment Retrieval Datasets}

\textbf{Charades-STA}~\cite{gao2017talltemporalactivitylocalization} is derived from the Charades dataset and contains 16,128 temporal annotations for moment retrieval. Each annotation consists of a natural language query describing a specific activity and the corresponding temporal boundary (start and end timestamps) within the video. The dataset focuses on daily indoor activities and requires fine-grained temporal understanding. Following standard convention, we report Recall@$k$ at IoU thresholds of 0.3, 0.5, and 0.7, as well as mean IoU (mIoU) on the test split.

\noindent \textbf{ActivityNet-Captions}~\cite{krishna2017densecaptioningeventsvideos} is a large-scale dataset for dense video captioning and temporal localization. It contains 20,000 videos with 100,000 temporally localized sentence descriptions. Videos are significantly longer than other benchmarks (averaging 120 seconds) and contain multiple events with temporal annotations. For moment retrieval evaluation, we use natural language queries to localize specific temporal segments. We report Recall@$k$ at IoU thresholds of 0.3, 0.5, and 0.7, and mIoU on the val-2 split following standard protocol.

\subsection{Model Prompts and Instructions}

We detail the specific prompts and instructions used across different components of VeRVE. All prompts are designed to be concise while clearly conveying the task objective to the model.

\subsubsection{Contrastive Learning Prompts (VeRVE-Embed)}

For training the embedding model with contrastive learning, we use task-specific prompts to generate unified embeddings:

\noindent\textbf{Video encoding:}
\begin{verbatim}
<video> Summarize this video in 
one word: <EOS>
\end{verbatim}

\noindent\textbf{Text encoding:}
\begin{verbatim}
<text> Summarize this text in 
one word: <EOS>
\end{verbatim}

\noindent\textbf{Image encoding (Stage 1):}
\begin{verbatim}
<image> Summarize this image in 
one word: <EOS>
\end{verbatim}

These prompts encourage the model to produce concise, semantically meaningful representations by focusing on the core content. The \verb|<EOS>| token's final hidden state serves as the embedding anchor, attending to the full multimodal context through causal attention.

\subsubsection{Re-ranking Prompts (VeRVE-Ranker)}

For the re-ranking stage, we formulate the task as a binary matching problem with the following prompt:

\noindent \textbf{Video-query matching:}
\begin{verbatim}
<video> <text> Does the text 
match the video? <EOS>
\end{verbatim}

Where \verb|<video>| represents the temporally ordered video frames and \verb|<text>| represents the query text. The model predicts a confidence score in $[0,1]$ from the \verb|<EOS>| token's hidden state via a linear projection head.

\subsubsection{Composed Query Prompts}

For composed video retrieval tasks (\textit{e.g.}, video+text$\rightarrow$video), we construct the query by concatenating multiple components with an explicit instruction:

\noindent \textbf{Composed query format:}
\begin{verbatim}
<source_video> <modification_text>
Encode the representation by 
considering the semantic change 
the source video would undergo 
under this modification: <EOS>
\end{verbatim}

For example:
\begin{verbatim}
<source_video> Switch this to a snowy 
mountain environment. Encode the 
representation by considering the 
semantic change the source video 
would undergo under this 
modification: <EOS>
\end{verbatim}

This instruction-based approach enables the model to jointly reason about the source visual content and the desired modification, producing a composed query embedding that captures the intended transformation.

\subsubsection{Moment Retrieval Processing}

For moment retrieval, we encode the query once and compute frame-level similarities. The query is encoded using the standard text encoding prompt:

\noindent \textbf{Temporal query encoding:}
\begin{verbatim}
<text> Summarize this text in one 
word: <EOS>
\end{verbatim}

Individual video frames are encoded separately using the image encoding prompt. No special temporal instructions are provided, as the model performs zero-shot localization through similarity-based peak detection over the temporal dimension.

\subsubsection{System Instruction}

We use different system instructions for each component of VeRVE to align with their specific objectives:

\noindent \textbf{VeRVE-Embed system prompt:}
\begin{verbatim}
You are a helpful assistant.
\end{verbatim}

\noindent \textbf{VeRVE-Ranker system prompt:}
\begin{verbatim}
You are a strict video text 
matching judge.
\end{verbatim}

For VeRVE-Embed, we use the standard Qwen system instruction to maintain consistency with the model's pre-training and general-purpose embedding generation. For VeRVE-Ranker, we employ a task-specific system prompt that emphasizes the discriminative nature of the matching task, encouraging the model to provide precise relevance assessments. These system instructions remain constant throughout training and inference for their respective components and precede all task-specific prompts described in the following sections.

\section{Baselines}

\subsection{Socratic Baseline}

To assess the effectiveness of our cross-modal contrastive training, we implement a caption-based retrieval baseline that approximates the zero-shot performance of the base MLLM without retrieval-specific training. \\

\noindent{\textbf{Method.} For each video, we first ask Qwen 2.5-VL to generate a detailed caption. We then embed both the generated captions and queries using GRIT-LM~7B~\cite{muennighoff2025generativerepresentationalinstructiontuning}, a state-of-the-art text embedding model, and perform retrieval by computing cosine similarities in the text embedding space. We use the following prompt for GRIT-LM:}

\begin{verbatim}
<caption text> Given a video caption, 
retrieve the most relevant video
\end{verbatim}

\textbf{Results.} This baseline achieves 32.6\% R@1 on MSR-VTT, substantially lower than VeRVE-Embed's 46.8\% R@1. The 14.2 percentage point gap demonstrates that direct cross-modal contrastive learning is essential for effective video-text retrieval, as caption-mediated approaches suffer from information loss and lack of query-specific adaptation.

\subsection{Other Baseline Methods}

We compare VeRVE against several state-of-the-art video retrieval methods, categorized by their retrieval architecture and training data scale.

\textbf{Embedding-only models.} VLM2Vec~\cite{jiang2025vlm2vectrainingvisionlanguagemodels}, VLM2Vec-V2\cite{meng2025vlm2vecv2advancingmultimodalembedding}, GVE \cite{guo2025universalvideoretrievalgeneralizing}, InternVideo2-Embed \cite{wang2024internvideo2scalingfoundationmodels} and CaRe~\cite{xu2025carebenchfinegrainedbenchmarkvideo} perform retrieval solely through embedding-based similarity search. VLM2Vec trains on diverse multi-task data mixtures spanning ${\sim}662K$ image-text pairs , while CaRe employs a two-phase approach: fine-grained video-caption alignment followed by retrieval adaptation on text-text pairs. VLM2Vec-V2, InternVideo2-Embed and GVE incorporate a significant amount of video-text data during training. InternVideo2-Embed refers to the embedding-only performance of the model with its vision-text matching head disabled as reported in \cite{guo2025universalvideoretrievalgeneralizing}.

\textbf{Joint-processing models.} InternVideo2~\cite{wang2024internvideo2scalingfoundationmodels} and LamRA~\cite{liu2024lamralargemultimodalmodel} combine embedding-based retrieval with re-ranking. InternVideo2, a specialized video foundation model, is trained on ${\sim}400$M video-image-audio-text samples with a learned image-text matching module (joint re-ranking style) and employs the dual-softmax scoring strategy. LamRA adapts MLLMs through multi-task instruction tuning and employs a re-ranker that generates ``Yes/No" text responses for relevance assessment.

In contrast, VeRVE achieves competitive performance with only ${\sim}700$K training samples (595K image-text + 105K video-text) through our focused two-stage contrastive training strategy and preference-based re-ranking objective.

\section{Composed Video Retrieval}

\subsection{Ablation Studies}

We conduct ablation experiments on CoVR-2 to understand the factors enabling zero-shot composed video retrieval. Results are presented in Table~\ref{tab:covr_ablation}.

\textbf{Input ordering.} Our default formulation (video first, then modification text) achieves 55.49\% R@1, while reversed ordering (text first, then video) drops to 49.64\% R@1 (-5.85 points). This degradation can be attributed to two factors: (1)~\textit{causal attention}, where the modification text can attend to the video in our formulation but not vice versa in the reversed case, limiting cross-modal reasoning; and (2)~\textit{training consistency}, as the model is trained with video-first ordering in all video-text pairs, making the reversed ordering a distribution shift at inference.

\textbf{Edit text importance.} Ablating the modification text and using only the video with a standard summarization prompt yields 45.15\% R@1, a 10.34 point drop. This validates that the model genuinely performs compositional reasoning by integrating both modalities, rather than simply retrieving based on source video similarity alone.

These ablations confirm that \textit{VeRVE-Embed}'s zero-shot composed retrieval capability emerges from effective joint encoding and leveraging the base MLLM's multimodal reasoning abilities.

\subsection{Training on CoVR Data}

While \textit{VeRVE-Embed} achieves strong zero-shot performance on composed video retrieval, we also evaluate its performance when directly trained on the CoVR training set using our contrastive learning framework. Results are shown in Table~\ref{tab:covr_trained}.

We train \textit{VeRVE-Embed$^\dagger$} on the CoVR training set for 1 epoch using the same two-stage contrastive training strategy and hyperparameters described in the main paper. The model is initialized from our video-text pre-trained checkpoint (Stage 2) and fine-tuned on CoVR video-text-video triplets, treating the composed query (source video + modification text) as the query and the target video as the positive candidate.

Our trained model achieves 68.3\% R@1, outperforming prior methods: +8.18 points over Thawakar et al.~\cite{thawakar2024composedvideoretrievalenriched} and +15.17 points over CoVR-BLIP~\cite{Ventura_2024}. This demonstrates that our contrastive learning framework effectively adapts to composed retrieval when provided with task-specific training data. The ${\sim}8$\% improvement over both zero-shot (55.49\%) and supervised baselines (60.12\%) validates the effectiveness of our training strategy for complex multimodal composition tasks.
\label{sec:covr_ablations}

\begin{table}[t]
\centering
\caption{Ablations on CoVR zero-shot composed video retrieval. We compare our standard formulation against variants that modify input ordering or remove the edit text.}
\label{tab:covr_ablation}
\vspace{-0.3cm}
\small
\scalebox{0.78}{
\begin{tabular}{p{3.0cm}|ccc|ccc}
\toprule
\multirow{2}{*}{\textbf{Ablation Setting}} 
& \multicolumn{3}{c|}{\textbf{T→V}} 
& \multicolumn{3}{c}{\textbf{V→T}} \\
\cmidrule(lr){2-4} \cmidrule(lr){5-7}
& \textbf{R@1} & \textbf{R@5} & \textbf{R@10} 
& \textbf{R@1} & \textbf{R@5} & \textbf{R@10} \\
\midrule
Default setup 
& 55.49 & 79.41 & 86.26 
& 56.05 & 79.85 & 86.86 \\
Reverse order 
& 49.64 & 74.60 & 81.97 
& 50.52 & 75.12 & 83.17 \\
Without edit text 
& 45.15 & 70.03 & 77.68 
& 45.07 & 70.39 & 78.45 \\
\bottomrule
\end{tabular}
}
\end{table}

\begin{table}[t]
\centering
\caption{Text + Video → Video Retrieval on COVR test set with \textit{VeRVE-Embed$^\dagger$} model trained constrastively on CoVR dataset}
\label{tab:covr_trained}
\begin{tabular}{lccc}
\toprule
\textbf{Model Name} & \textbf{R@1} & \textbf{R@5} & \textbf{R@10} \\
\midrule
COVR-BLIP \cite{Ventura_2024}  & 53.13& 79.93 & 86.85 \\
Thawakar et. al \cite{thawakar2024composedvideoretrievalenriched} & 60.12 & 84.32 & 91.27 \\
\textit{VeRVE-Embed$^\dagger$ 7B (Ours)} & \textbf{68.3} & \textbf{88.6} & \textbf{93.4} \\
\bottomrule
\end{tabular}
\end{table}

\end{document}